\documentclass[peerreview]{IEEEtran}
\usepackage{cite} 
\usepackage{url} 
\usepackage[utf8]{inputenc} 
\usepackage{graphicx}
\usepackage{amsmath}
\usepackage{amssymb}
\usepackage{booktabs}
\usepackage{multirow}
\usepackage{enumitem}
\usepackage{subfloat}
\usepackage{arydshln}

\usepackage{algorithm}
\usepackage{algorithmic}

\usepackage[capitalize]{cleveref}
\crefname{section}{Sec.}{Secs.}
\Crefname{section}{Section}{Sections}
\Crefname{table}{Table}{Tables}

\begin{document}
\title{Progressive Multi-stage Interactive Training in Mobile Network for Fine-grained Recognition}



\author{Zhenxin Wu, \and  Qingliang Chen, \and Yifeng Liu, \and Yinqi Zhang, \and Chengkai Zhu, \and Yang Yu

{2533653096@qq.com}
}


\maketitle
\tableofcontents
\listoffigures
\listoftables

\IEEEpeerreviewmaketitle
\begin{abstract}
Fine-grained Visual Classification (FGVC) aims to identify objects from subcategories. It is a very challenging task because of the subtle inter-class differences. Existing research applies large-scale convolutional neural networks or visual transformers as the feature extractor, which is extremely computationally expensive. In fact, real-world scenarios of fine-grained recognition often require a more lightweight mobile network that can be utilized offline. However, the fundamental mobile network feature extraction capability is weaker than large-scale models. In this paper, based on the lightweight MobilenetV2, we propose a Progressive Multi-Stage Interactive training method with a Recursive Mosaic Generator (RMG-PMSI). First, we propose a Recursive Mosaic Generator (RMG) that generates images with different granularities in different phases. Then, the features of different stages pass through a Multi-Stage Interaction (MSI) module, which strengthens and complements the corresponding features of different stages. Finally, using the progressive training (P), the features extracted by the model in different stages can be fully utilized and fused with each other. Experiments on three prestigious fine-grained benchmarks show that RMG-PMSI can significantly improve the performance with good robustness and transferability.
\end{abstract}

\section{Introduction}

\label{sec:intro}
Fine-grained Classification aims to identify different subcategories of the same category, e.g., different kinds of birds, cars or planes. Early work mainly used the method of strong supervision in which people manually marked specific regions \cite{DBLP:conf/iccv/XieTHYZ13, DBLP:conf/eccv/ZhangDGD14, DBLP:journals/corr/BransonHBP14,DBLP:conf/cvpr/HuangXTZ16, DBLP:journals/corr/WeiXW16, DBLP:conf/cvpr/BergB13}. These methods required a lot of manpower and were prone to errors, which led to performance degradation \cite{DBLP:conf/iccvw/Krause0DF13}. Therefore, this field of research has gradually shifted to weak supervision approaches that do not require explicit labeling of regions \cite{DBLP:conf/iccv/ZhengFML17, DBLP:conf/iccvw/Krause0DF13, DBLP:conf/eccv/YangLWHGW18, DBLP:conf/cvpr/ChenBZM19, DBLP:conf/aaai/ZhuangW020,DBLP:conf/eccv/DuCBXMSG20, DBLP:journals/corr/abs-2103-07976}. These efforts incorporate features from different regions or pair of interactive comparative learning to make features more discriminative, resulting in significant performance improvements.

However, the existing approaches mentioned above mainly focus on using large-scale CNN \cite{DBLP:journals/corr/SimonyanZ14a, DBLP:conf/cvpr/HeZRS16, DBLP:conf/cvpr/XieGDTH17, DBLP:conf/cvpr/HuangLMW17} or visual transformer \cite{DBLP:conf/iclr/DosovitskiyB0WZ21} as feature extractor, which is well-known to be computationally expensive. In fact, fine-grained classification  scenarios often require a more lightweight mobile network that can be applied offline. For example, ornithologists conducting field research need to be able to use mobile devices to quickly identify birds they find. In the field of intelligent traffic, traffic police also rely on mobile devices to quickly identify vehicle models when performing tasks. However, the large-scale CNNs has a hefty price tag and makes it difficult to recognize objects in real-time. Therefore, we need new mobile networks to meet the requirements for lightweight models and fast offline recognition in the scenarios mentioned above.

\begin{figure}
  \centering
   \includegraphics[width=1\linewidth]{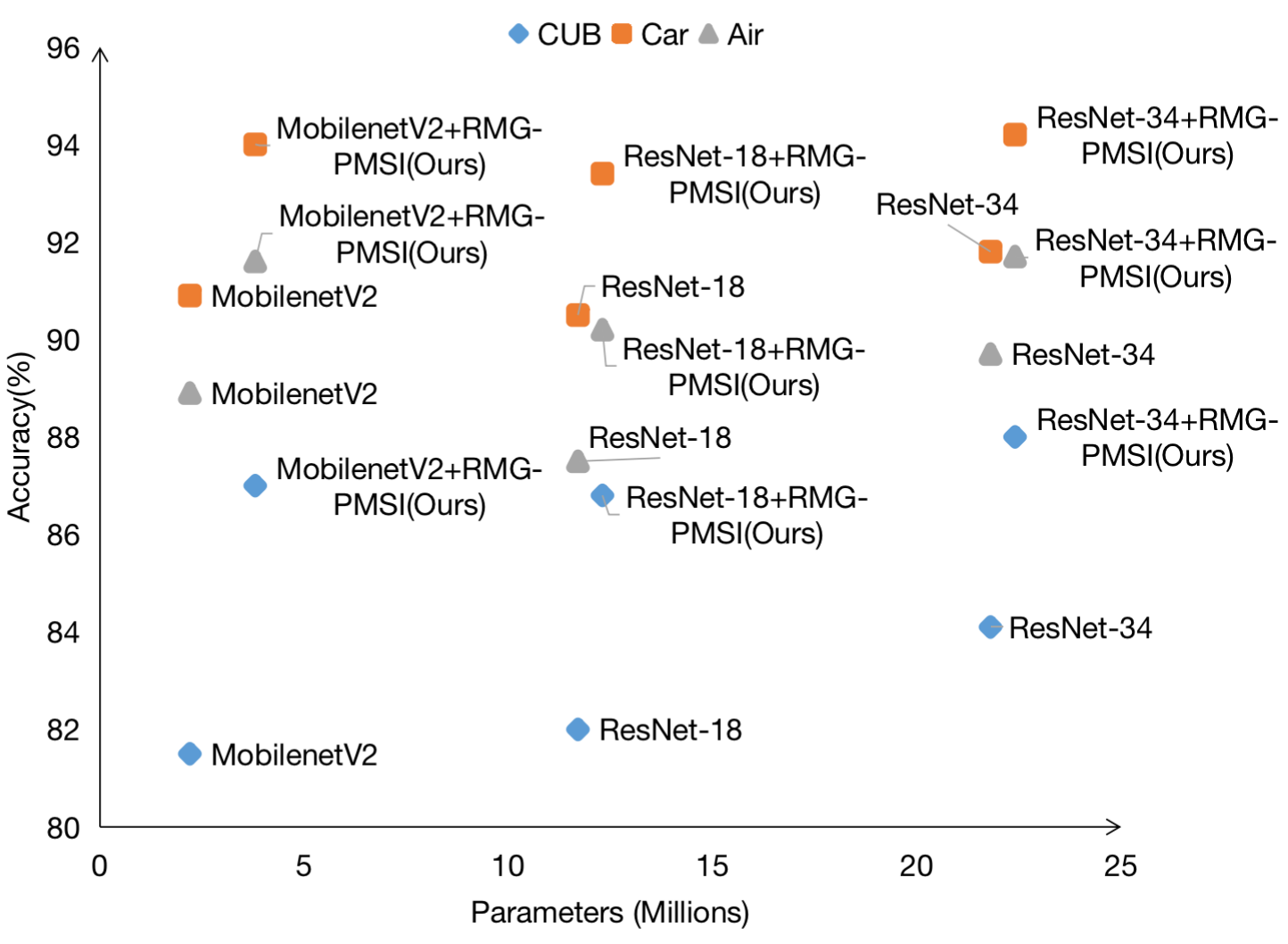}
   \caption{An overview of performance comparison of baseline and RMG-PMSI in CNN backbones on three datasets.}
   \label{fig:view}
\end{figure}

At present, because the feature extraction capability of lightweight mobile network \cite{DBLP:conf/cvpr/SandlerHZZC18,DBLP:conf/eccv/ZhouHCFY20} is weaker than the large-scale model, there is no research on a mobile network specifically for fine-grained recognition. However, if we can improve the utilization of the features based on the limited feature extraction capability, the performance of the model will be enhanced. Intuitively, the shallow layer of an end-to-end CNN will extract fine-grained local features of the image. As the network depth increases, the receptive field also expands. CNN will gradually extract the global features and use them for classification. During the process, much of the local information at the shallow layer that is helpful for classification is not taken into account at the end. So, if we can fully combine all of the local and deep global features that are extracted from different stages of CNN, the features of different stages can jointly boost the classification accuracy. As a result, the feature extraction ability of the model can be fully utilized to improve the performance of the lightweight model.

Inspired by the idea above, and based on the lightweight MobilenetV2, we propose a Progressive Multi-Stage Interactive training method for lightweight mobile networks using a Recursive Mosaic Generator (RMG-PMSI). Firstly, we introduce a Recursive Mosaic Generator (RMG) that generates images containing different granularities at different stages. Then, the features of different stages go through a Multi-Stage Interactive module (MSI) to reinforce and supplement the corresponding features of different stages. Finally, using the progressive training (P), the model focuses on learning stable local fine-grained features in the shallower layer and more abstract and large-grained global features in the deeper layer in the next phase. In addition, during the training process of each phase, different stages produce an interactive supplement instead of being disassociated, as it ensures the consistency in the whole network training process. By using RMG-PMSI as the training mode, the features extracted from the model at different stages can interact and be complementary with each other, which significantly improves the model performance.

The proposed method has been extensively evaluated on three prestigious fine-grained visual classification benchmarks (CUB-200-2011\cite{WahCUB_200_2011}, Stanford Cars \cite{DBLP:conf/iccvw/Krause0DF13}, Aircraft\cite{DBLP:journals/corr/MajiRKBV13}). An overview of the performance comparison can be seen in \cref{fig:view}. Compared to the baselines, the RMG-PMSI method provides a significant performance improvement. At the same time, we conducted two experiments testing robustness. The results show that the RMG-PMSI has good transferability to models of different scales and can equip them with significantly more anti-interference power. In summary, the main contributions of this paper are as follows:
\begin{itemize}

\item  A new data augmentation method for FGVC named Recursive Mosaic Generator (RMG) is introduced that can help models focus on features of different granularities at different training phases.

\item  A novel Progressive Multi-Stage Interactive training method (PMSI) for lightweight mobile networks is proposed for Fine-grained Visual Classification (FGVC), which can make better utilization of features and strengthen the anti-interference ability substantially.

\item The proposed RMG-PMSI approach has been implemented and evaluated on standard benchmarks, with performance being significantly improved, and moreover with good robustness and transferability, leading to the potential application on mobile devices.
    
\end{itemize}

\section{Related Work}
\label{sec:related work}
In this section, we mainly discuss methods related to FGVC, data argumentation, multi-stage feature fusion and progressive training.
\subsection{Fine-grained visual classification}
As visual models continue to evolve, the research on FGVC has shifted from strongly supervised methods with additional bounding boxes \cite{DBLP:conf/cvpr/BergB13,DBLP:conf/iccv/XieTHYZ13,DBLP:conf/eccv/ZhangDGD14,DBLP:conf/cvpr/HuangXTZ16} to weakly supervised ones with category labels only \cite{DBLP:conf/cvpr/FuZM17,DBLP:conf/iccv/ZhengFML17, DBLP:conf/eccv/YangLWHGW18, DBLP:conf/iccv/WeiY0DL19, DBLP:conf/cvpr/GeLY19, DBLP:conf/icmcs/ZhengC0M20, DBLP:journals/tvt/LiYCMC19}. Most of the weakly supervised methods aim at locating the most discriminative regions in the image. For example, Fu et al. \cite{DBLP:conf/cvpr/FuZM17} found that region detection and fine-grained feature learning could reinforce each other and constructed a series of networks during prediction to locate more differentiated regions for the following networks. Lin et al.\cite{DBLP:conf/iccv/LinRM15} presented bilinear pooling on the representation of two local patches in the image to learn more representative features. However, these approaches used a large-scale framework to extract high-level abstraction features at the end of an end-to-end network without utilizing shallow local fine-grained features. Due to the mobile network's limited feature extraction capability, the full utilization of features extracted at different stages remains to be explored.

\begin{figure*}
  \centering
   \includegraphics[width=1\linewidth]{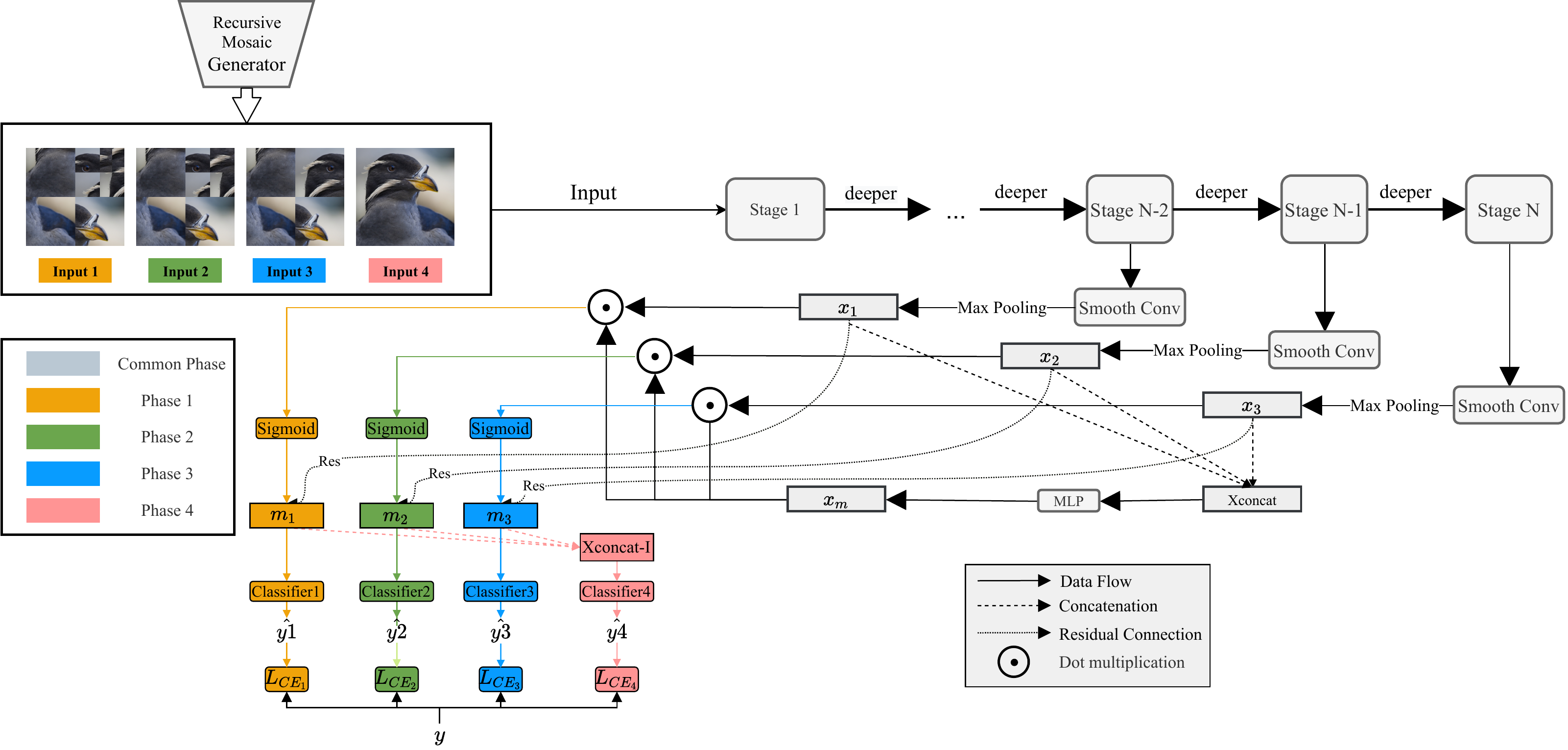}
   \caption{The structure of PMSI is illustrated. There are $StageNum+1$ phases at each iteration. Here, we set $StageNum=3$ for explanation. We set the increase in the number of feature map channels as the division between this stage and the next stage. There are a total of $N$ stages in the backbone. The Smooth Conv Block consists of two convolutional layers which are used to convert feature maps with different channel numbers generated by different stages into same number of channels. The MLP (multi-layer perceptron) has two fully connected layers. The classifier contains two fully connected layers with a softmax layer. The input image of each phase is different (corresponding to input 1 to input 4). Each phase produces different classification results and the common parts of each phase are marked in gray in the figure.}
   \label{fig:PMSI_overview}
\end{figure*}

\subsection{Data argumentation for FGVC}
In the field of computer vision, there are many popular data augmentation methods. For example, for conventional image recognition, data augmentation based on fusion of different images is widely used \cite{DBLP:conf/iccv/YunHCOYC19, DBLP:conf/iclr/ZhangCDL18}. Based on these tricks, a trainable image fusion augmentation is adopted to achieve great performance improvement in end-to-end training for FGVC \cite{DBLP:conf/aaai/HuangWT21}. In a weak supervision network, Jigsaw Puzzle is often used as an initialization method to achieve better conversion performance \cite{DBLP:conf/cvpr/WeiXRXS00Y19}. In a recent study, PMG \cite{DBLP:conf/eccv/DuCBXMSG20} encouraged the generation of different granular inputs through the Jigsaw Puzzle at different stages. However, none of these methods explicitly generated fine-grained local features and large-grained global features in a single image, which are very substantial  for FGVC. In this paper, we address the deficit, and propose a Recursive Mosaic Generator (RMG) accordingly by combining the mosaic augmentation and recursion.

\subsection{Multi-stage feature fusion and progressive training}

Multi-stage feature fusion is often applied in the field of object detection \cite{DBLP:conf/cvpr/LinDGHHB17,DBLP:journals/corr/abs-1712-00960,DBLP:conf/cvpr/RedmonF17}. By combining information from different stages, we can better distinguish the background and objects. In FGVC, API \cite{DBLP:conf/aaai/ZhuangW020} can well solve the problem of image confusion due to the nuances between fine-grained image classes by fusing two similar fine-grained image features to find distinguishing clues.

Progressive training is widely used in generation tasks  \cite{DBLP:conf/cvpr/KarrasLA19,DBLP:conf/iccv/ShahamDM19,DBLP:conf/cvpr/AhnKS18}, which starts with low resolution images and gradually improves the resolution by adding new layers to the network. Rather than learning features from all scales, the strategy allows the network to discover the large-scale structures of the image distribution and shift its attention to finer-scale details. In FGVC, \cite{DBLP:conf/eccv/DuCBXMSG20} adopted a progressive training method for image classification, which guided the training of the next stage by the training of the previous stage, gradually shifting the focus from local features to global features.

In this paper, by combining multi-stage feature fusion and progressive training, we propose a progressive multi-stage interactive training approach, which can make features of different stages jointly cooperate in the training and ensure the conformity of training goals. Together with the proposed RMG trick, the feature extraction capability of the model is fully utilized, and the model performance can thus be significantly improved.

\section{The Proposed Approach}
\label{sec:method}
In this section, we will present our proposed approach of RMG-PMSI, integrating Progressive Multi-Stage Interactive training (PMSI) and Recursive Mosaic Generator (RMG). The main idea is to enable interactive learning of features at different stages in the training phase, which can make better use of features of different granularities extracted from the network to recognize fine-grained images and improve the robustness of the model. 

\subsection{Overview}
PMSI consists of two key components: (i) Recursive Mosaic Generator (RMG) for generating multi-granularities inputs; (ii) Progressive Multi-Stage Interaction (PMSI), as shown in \cref{fig:PMSI_overview}.

To be more specific, firstly, a batch of fine-grained images is fed into the RMG to generate inputs of different phases. The granularities of the input images in different phases are different so that each phase focuses on learning features from fine-grained granularities to larger-grained ones. After the input goes through the backbone, the features generated by multiple stages are extracted respectively. Then, the input of different stages generates interaction vectors $x_{m}$, and through a gate mechanism, the corresponding features of different stages are strengthened and supplemented. Finally, in each training phase, the network focuses on training the information of the corresponding granularity extracted by one of the stages. Through the above process, the model can be guided to learn stable fine-grained information in the shallow layer. With the progress of training, the model focuses on learning more abstract and larger-grained global information in the deeper layer in the next phase. In addition, in the training process of each phase, different stages are supplemented interactively rather than being separated and disassociated, which ensures the consistency of the goal in the whole network training phase, filtering out the disturbance of other stage features caused by the independent training of different stages.

\subsection{Recursive Mosaic Generator}
Mosaic data augmentation~\cite{DBLP:journals/corr/abs-2004-10934} is an effective mean of boosting performance in object detection. Meanwhile, Jigsaw Puzzle solving~\cite{DBLP:conf/cvpr/WeiXRXS00Y19,DBLP:conf/eccv/DuCBXMSG20} was found to be effective in self-supervised learning and progressive training tasks. We get inspiration from the above two techniques and propose accordingly a Recursive Mosaic Generator (RMG) for a picture to adapt to our progressive multi-stage interactive training.
The goal of the RMG is to design regions of different granularities and force the network to learn information specific to the corresponding level of granularity in different training phases. The recursive idea is that the generator will randomly select one of the patches based on the mosaic image generated in the previous step and perform the mosaic operation again. Then, the image contains a large granularity of global information that is difficult to learn, and furthermore, the stable and fine-grained local information that is easy to learn.

Expressly, assume that the input image $p\in W\times H\times C$ is randomly divided into $2\times2$ patches in the first recursion, and the size of each patch is $\frac{1}{2}W\times\frac{1}{2}H\times C$. In the second recursion, we will randomly divide one of the four patches generated in the first recursion into $2\times2$ patches, while the remaining three patches are unchanged. Similarly, one of the patches generated based on the previous recursion will be randomly selected to enter the following recursion in each recursion. Finally, we combine all the generated patches from the recursion into a new graph $G\left(p,r\right)$. Here, the number of recursions - the granularity contained in the image - is controlled by the hyperparameters $r$.

For the selection of $r$ in each stage, it is necessary to meet the following requirements. The size of the smallest patch in the images generated by RMG should be smaller than the receptive field of the corresponding stage; otherwise, it will increase the difficulty of learning in the corresponding stage of the network and affect the model's performance. Therefore, the size of the patch should increase proportionally with the size of the receptive field in each stage. For the input of the last phase, we assume $r=0$ (i.e., the initial image). So for each previous stage, the recursive hyperparameter $r$ should be increased by one. Assuming that phase \_num$=4$, the input image for phase 1 is $G\left(p,r=3\right)$.

\begin{figure}
  \centering
   \includegraphics[width=1\linewidth]{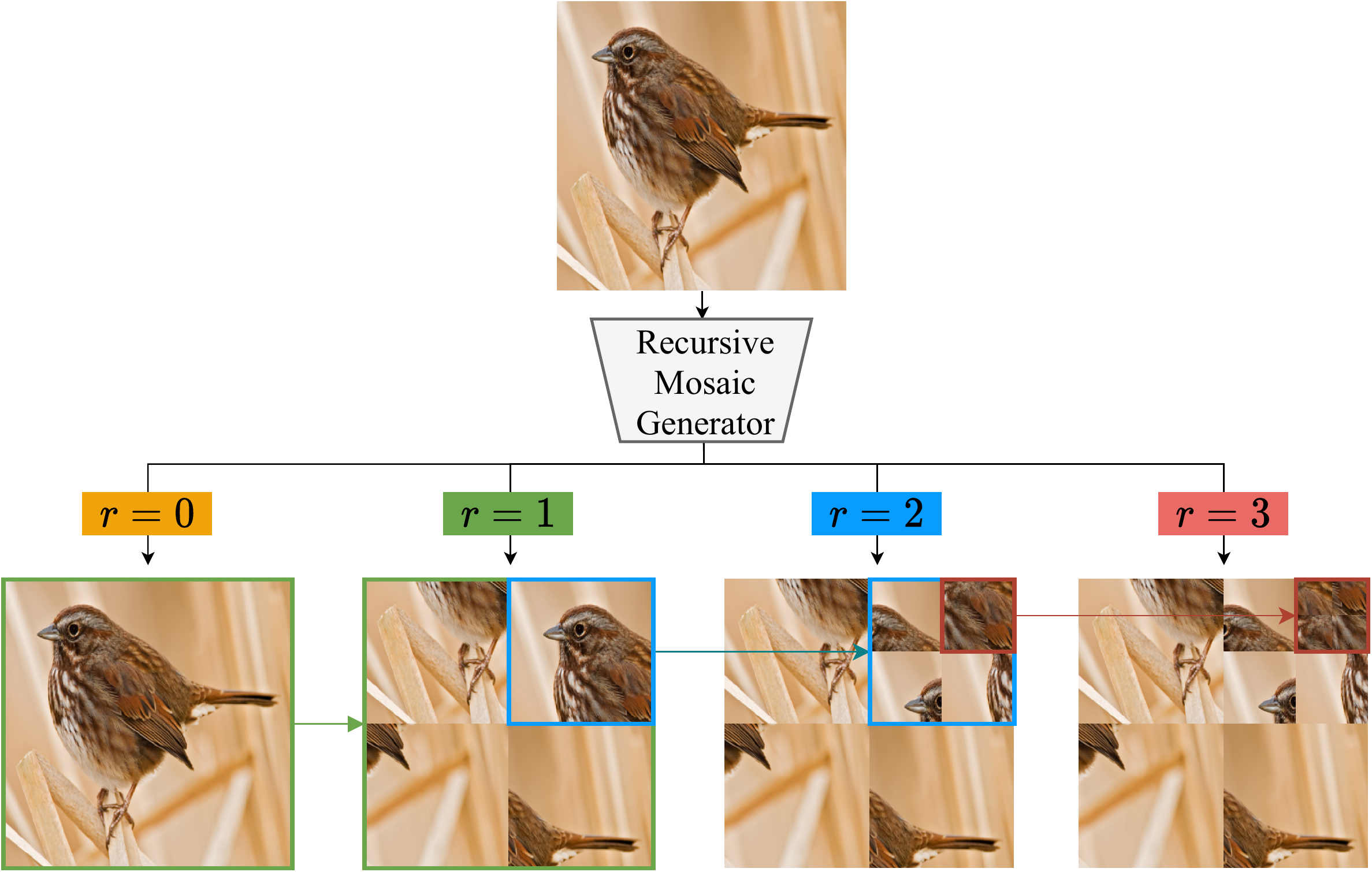}
   \caption{Recursive Mosaic Generator (RMG). In each phase, $2\times2$ mosic operation is performed again based on a random patch from the previous phase.}
   \label{fig:RMG}
\end{figure}

As shown in~\cref{fig:RMG}, we use the RMG to process the initial image and obtain new images as the input to different phases of our network, and the new image carries the same label $y$ as the initial image. It is worth emphasizing that the RMG does not always guarantee that all patch parts contain meaningful information when $r$ increases. Therefore, the number of recursions should be controlled within a reasonable range (i.e., $r\le3$). Otherwise, if the patch is too small, too much noise will be introduced, making it difficult to learn.

\subsection{Progressive multi-stage interaction}
In this section, we will introduce the Progressive Multi-stage Interaction (PMSI) module, which is the essential part of our proposed model. PMSI can be divided into two parts of multi-stage interaction and progressive training.
\subsubsection{Multi-stage interaction}
As shown in~\cref{fig:PMSI_overview}, the backbone generates different output feature maps in different stages. Then, the feature map passes through Smooth Conv and global max pooling to generate feature vectors $x_n\in\mathbb{R}^c$ , $n\in\left\{N-StageNum+1,\cdots,N-1,N\right\}$ at different stages, where $c$ is the number of channels of feature vectors. Then, we learn a mutual vector $x_m$ from $x_{N-StageNum+1}$ to $x_N$ as follows:

\begin{equation}
x_m=f_m\left(f_c(x_{N-StageNum+1},\cdots,x_{N-1},x_N\right)).
\end{equation}

where $f_c$ is a concatenation operation and $f_m$ is a mapping funtion of $\left[x_{N-StageNum+1},\cdots,x_{N-1},x_N\right].$ Specifically, we use the MLP as the mapping function. Since $x_m\in\mathbb{R}^c$ is adaptively summarized from features of different stages, it often contains information of different granularities.

After learning the mutual vector $x_m$, we proceed to expand it to $x_{N-StageNum+1},\cdots,x_{N-1},x_N$. The main reason is that, by interacting the mutual vector with the feature vectors of different stages, the features generated from each stage can be supplemented with different granularities from other stages, so as to make better advantage of features of different levels. In particular, we perform channel-wise product between $x_m$ and $x_n$. Then, we add a sigmoid function $\sigma$ to generate supplemental gate vectors $g_n\in\mathbb{R}^c$ corresponding to different stages as follows:

\begin{equation}
g_n=\sigma\left(x_m\odot x_n\right).
\end{equation}

where $n\in\left\{N-StageNum+1,\cdots,N-1,N\right\}$. Finally, we introduce an interactive supplement mechanism for the features extracted from each stage via residual connections as follows:

\begin{equation}
m_n=x_n+g_n\odot x_n.
\end{equation}

where $n\in\left\{N-StageNum+1,\cdots,N-1,N\right\}$, and $m_n\in\mathbb{R}^c$ is the output of multi-stage interactive module.

\subsubsection{Progressive training strategy} 
The proposed RMG-PMSI model adopts the progressive training strategy to effectively capture discriminative information of different granularities. It could be implemented on any popular CNN feature extractor, such as the MobilenetV2 \cite{DBLP:conf/cvpr/SandlerHZZC18} and ResNet \cite{DBLP:conf/cvpr/HeZRS16}. 

The model gets the feature vector $m_n$ after multi-stage interactions. Thereafter, a classification module $F_{classifier}^n$ consisting of two fully connected layers with batch  normalization \cite{DBLP:conf/icml/IoffeS15} and the activation function ELU \cite{DBLP:journals/corr/ClevertUH15} corresponding to the $n$-th stage, predicts the probability distribution over the classes as 

\begin{equation}
{\hat{y}}_n=F_{classifier}^n\left(m_n\right). 
\end{equation}

Here, we consider the last $StageNum$ stages with $n\in\left\{N-StageNum+1,\cdots,N-1,N\right\}$. Finally, we concatenate the outputs from the last $StageNum$ stages as
\begin{equation}
m_{concat}=\mathrm{concat}\left[m_{N-StageNum+1},\cdots,m_{N-1},m_N\right].
\end{equation}

This is followed by an additional classification module
\begin{equation}
{\hat{y}}_{concat}=F_{classifier}^{concat}\left(m_{concat}\right).
\end{equation}

During training, each iteration contains $StageNum+1$ phases. For training, we compute the cross entropy loss $L_{CE}$ between the ground truth label $y$ and the predicted output from each phase. At each iteration, a batch of data $d$ is used for $StageNum+1$ phases, and we only focus on training one stage’s output $y_n$ at each phase. It needs to be emphasized that, except for the classifier, all parameters that are used in the current phase prediction will be optimized, even they might have been updated in previous phases, and the strategy helps all stages in the model collaborate together. The detailed implementation of the training algorithm of our RMG-PMSI model is shown in~\cref{alg:2}.

\begin{algorithm}
    \caption{RMG-PMSI}
    \label{alg:2}
    \begin{algorithmic}
        \STATE {Training data set $D$, \\ Training data for a batch $x$,\\ Training label for a batch $y$. }
        \FOR{$epoch$ $ \in [0,epochs)$}
            \FOR{$b$ $ \in [0,D/batchSize)$}
                \STATE {$x,y \Leftarrow$ batch \ $b$ \ of\ $D$}
                
                \FOR{$n  \in$ $[N - StageNum + 1, N]$}
                \STATE {$r \Leftarrow$ N - $n+1$}
                \STATE {$X \Leftarrow$ $G(x,r)$}
                \STATE {$m_{n} \Leftarrow F_{n}^{e}(X)$ }
                \STATE{$\hat{y}_{n}\Leftarrow F_{classifier}^{n}(m_{n})$}
                \STATE {Backprop$(L_{CE}(\hat{y}_n,y))$}
                \ENDFOR
            \STATE{$m_{concat}\Leftarrow$ $concat[m_{N-StageNum+1},...,m_{N-1},m_{N}]$}
            {$\hat{y}\Leftarrow F_{classifier}^{concat}(m_{concat})$}
            \STATE{Backprop $(L_{CE}(\hat{y_{concat},y})$}
            \ENDFOR
        \ENDFOR
    \end{algorithmic}
\end{algorithm}

In the testing phase, if we only use $\hat{y}_{concat}$  as the prediction, the final result of our module can be expressed as 

\begin{equation}
P_{concat}=argmax(\hat{y}_{concat}).
\end{equation}

Nevertheless, the predictions of each phase are complementary. Hence, we combine all outputs to get the final result which can be calculated as

\begin{equation}
P_{mix}=argmax \left ((\sum_{n=N-StageNum+1}^{N} \hat{y}_{n})+\hat{y}_{concat} \right).
\end{equation}

\section{Experiments}
\label{sec:experiments}
In this section, we will introduce the datasets and implementation details of the experimental studies. Firstly, a series of ablation studies has been conducted to demonstrate the contributions of each module to the performance of our model. Next, we compare our model's performance to other state-of-the-art fine-grained classification counterparts. Moreover, we have carried out two experiments to test the robustness of RMG-PMSI. Finally, we visualize and qualitatively analyze the interior of the model.
\begin{table}[H]
\caption{Statistics of benchmark datasets}
\label{table:1}
\centering
\begin{tabular}{cccc}
\toprule
Dataset        & \#Classes & \#Train & \#Test \\ 

\toprule
CUB-200-2011   & 200       & 5994    & 5794   \\
Standford Cars & 196       & 8144    & 8041   \\
FGVC Aircraft  & 100       & 6667    & 3333   \\ 
\toprule
\end{tabular}
\end{table}
\subsection{Dataset and Implementation Details}
We have evaluated the performance of the proposed method on three prestigious fine-grained benchmarks: CUB-200-2011 (CUB) \cite{WahCUB_200_2011}, Stanford Cars (Car) \cite{DBLP:conf/iccvw/Krause0DF13} and FGVC-Aircraft (Air) \cite{DBLP:journals/corr/MajiRKBV13}, with details shown in~\cref{table:1}. It is worth emphasizing that in all experiments, the category labels of the images are the only annotations used for training.

We run all the experiments on the GTX 2080Ti GPU cluster using PyTorch with a version higher than 1.8.0. The method we proposed was evaluated on the widely used mobile network of MobilenetV2. During the training phase, we adjusted the input image to 512 × 512 and randomly cropped it to 448 × 448 after a random horizontal flip. During the test, we resized the image to 512 × 512 and cropped it from the center to 448 × 448 as input. We used Stochastic Gradient Descent (SGD) to optimize our network and we applied the pre-trained model on ImageNet. For pre-trained convolution layers, the initial learning rate was 0.0001 and reduced by the cosine annealing schedule. For newly added layers, the initial learning rate was 0.001. For all the aforementioned models, we trained them for up to 150 epochs with batch size as 32 and used a weight decay as 0.0005 and a momentum of 0.9. Besides, during training phase, we froze the backbone and only trained the newly-added layers in the first 5 epochs.

\subsection{Ablation studies}
\textbf{The impact of StageNum.} To demonstrate the effecacy of progressive interaction training, we conducted experiments without a Recursive Mosaic Generator (RMG) in CUB dataset.

We set the increase in the number of feature map channels as the division between this stage and the next stage. In order to obtain the best performance and prevent excessive noise being introduced by too shallow layer features, we use the last 5 stages as our experimental setup. The StageNum increases from 1 to 5. We use Top-1 Accuracy (Acc) as the evaluation criterion. The results of the experiment are shown in \cref{table:2}, where Concat is the result of the $P_{concat}$ (it is worth noting that when S={5}, Concat is equivalent to the concatenation of the output of stage 5 and its output after MLP) and Mix represents a mixed classification result $P_{mix}$. As the results show, when the number of stages (S) involved in interactive training is less than 4, the increase in StageNum improves the model's performance. Both Concat Accuracy and Mix Acc Accuracy began to decline when the StageNum=4. This might be caused by the low-stage layer focusing on class-independent features. However, the additional supervision forces the low-stage layer (S={1,2}) to focus prematurely on the features associated with classification, thus introducing too much low-stage noise to the high-stage classification through progressive interaction, resulting in a decrease in accuracy. In addition, using multi-stages of progressive interaction can lead to increased training costs. To sum up, we use the last three phases (S={5,4,3}, StageNum=3) as the optimal choice for progressive interactive training.
\begin{table*}[]
\caption{The performance of the proposed model when interacting at different stages.}
\label{table:2}
\centering
\begin{tabular}{lccccccc}
\toprule
\multicolumn{1}{c}{}  & \multicolumn{7}{c}{Acc(\%)}                                                                                                                                      \\ \toprule
Stage(S) / StageNum   & \multicolumn{1}{l}{S1} & \multicolumn{1}{l}{S2} & \multicolumn{1}{l}{S3} & \multicolumn{1}{l}{S4} & \multicolumn{1}{l}{S5} & \multicolumn{1}{l}{Concat} & Mix         \\
\{5\} / 1             & -                      & -                      & -                      & -                      & 84.2                   & 84.2                       & 84.3        \\
\{5, 4\} / 2          & -                      & -                      & -                      & 84.7                   & 84.5                   & 84.9                       & 85.2        \\
\{5, 4, 3\} / 3       & -                      & -                      & 82.9                   & 84.8                   & 84.3                   & \textbf{85.5}              & \textbf{85.9} \\
\{5, 4, 3, 2\} / 4    & -                      & 80.5                   & 83.3                   & 84.7                   & 83.9                   & 85.1                       & 85.7        \\
\{5, 4, 3, 2, 1\} / 5 & 74.2                   & 81.7                   & 83.0                     & 84.5                   & 84.3                   & 84.9                       & 85.1   \\ 
\toprule
\end{tabular}
\end{table*}

\textbf{The contribution of different components.} Specifically, RMG-PMSI can be split into three sub-modules, recursive mosaic generator (R), progressive training (P), and multi-stage interaction (M). To justify their contributions and joint effort, we validate all of their possible combinations in the CUB dataset\footnote{It should be noted beforehand that R must work with P.}. The results of the experiment are shown in~\cref{table:3}.

\begin{table}[]
\caption{The contribution of each component in RMG-PMSI.}
\centering
\label{table:3}
\begin{tabular}{lcc}
\toprule
\multicolumn{1}{c}{Stage} & Concat Acc (\%)  & Mix Acc (\%) \\ \toprule
Baseline                  & 81.5          & -                 \\
+ M                       & 85.1          & -                 \\
+ P                       & 85.2          & 85.5              \\
+ P \& M                  & 85.5          & 85.9              \\
+ P \& R                  & 86.4          & 86.8              \\ 
+ P \& M \& R             & \textbf{86.6} & \textbf{87.0}       \\ \toprule
\end{tabular}
\end{table}

We can see from the results that both multi-stage interaction (M) and progressive training (P) can significantly improve the model. Combining them further drives accuracy forward. This shows that progressive training (P) and multi-stage interaction (M) can effectively benefit rather than offset each other. Recursive mosaic generator (R) works very well with progressive training (P), helping the model pay more attention to the fine-grained local information in shallow and global features in deep stages. The above experiments demonstrate the power of the components of our RMG-PMSI model and their good collaboration.
\begin{table}[]
\caption{Accuracy of different data augmentations.}
\centering
\label{table:4}
\begin{tabular}{lcc}
\toprule
\multicolumn{1}{c}{data argumentation} & Concat Acc (\%)  & Mix Acc (\%) \\ \toprule
baseline                                 & 85.5          & 85.9                   \\
cutmix                                   & 85.1          & 85.3                   \\
mixup                                    & 85.6          & 86.0                   \\
snapmix                                  & 85.1          & 85.1                   \\
jigsaw                                   & 86.3          & 86.6                   \\ 
RMG (ours)      & \textbf{86.6} & \textbf{87.0}            \\ \toprule
\end{tabular}
\end{table}
\subsection{Comparative studies}
\textbf{Methods for augmenting fine-grained data.} We compared our RMG with other data augmentation methods that are widely used in fine-grained classifications on CUB200, including cutmix \cite{DBLP:conf/iccv/YunHCOYC19}, mixup \cite{DBLP:conf/iclr/ZhangCDL18}, snapmix \cite{DBLP:conf/aaai/HuangWT21}, and jigsaw \cite{DBLP:conf/eccv/DuCBXMSG20}. As can be seen in~\cref{table:4}, stitching different pictures does not bring much improvement or even decrease the model performance. We believe that it happens because progressive training and multi-stage interaction encourage the fusion of different features of an image from the local level to the global level. The stitching-based image augmentation greatly disturbs the process of learning features of different stages, which makes it difficult to boost the model performance.
On the other hand, the puzzle method jigsaw performs well. Compared to jigsaw, our method can make the image contain both fine-grained local features and large-grained global features, which promote better collaborative learning at different stages and thus achieve optimal performance.

\textbf{Models for Fine-grained recognition.} To further illustrate the advantages of our RMG-PMSI, we compare it with the widely used state-of-the-art fine-grained approaches on MobilenetV2 in CUB dataset. The results of the experiment are shown in~\cref{table:5}. From the table, we can see that because B-CNN, HBP, PC, and API rely more on the abstract features extracted by the model in deeper stages, their performances are limited by the feature extraction capability of the MobilenetV2. For PMG and RMG-PMSI, the performance of the model can be greatly improved. The key to improving performance on a lightweight mobile network is to take full advantage of the features that the model extracts at different stages. Finally, the RMG-PMSI offers better performance improvement than PMG because RMG and multi-stage interaction are better equipped to help different stages work together during the progressive training.

\begin{table}[]
\caption{Comparison with other state-of-the-art fine-grained recognition models}
\centering
\label{table:5}
\begin{tabular}{ccc}
\toprule
Models         & (Concat) Acc (\%)  & Mix Acc (\%) \\ \toprule
B-CNN \cite{DBLP:conf/iccv/LinRM15}    & 83.2          & -                      \\
HBP \cite{DBLP:conf/eccv/YuZZZY18}      & 85.1          & -                      \\
PC \cite{DBLP:conf/eccv/DubeyGGRFN18}      & 84.5          & -                      \\
API \cite{DBLP:conf/aaai/ZhuangW020}     & 85.7          & -                      \\
PMG \cite{DBLP:conf/eccv/DuCBXMSG20}    & 85.9          & 86.5                   \\ 
\multicolumn{1}{c}{\begin{tabular}[c]{@{}c@{}}RMG-PMSI\\ (ours)\end{tabular}} & \textbf{86.6} & \textbf{87.0}            \\ \toprule
\end{tabular}
\end{table}

\begin{table}[]
\caption{The results on other stronger backbones}
\resizebox{1.05\linewidth}{!}{
\centering
\label{table:6}
\begin{tabular}{c|ccc}
\hline \hline
Backbone & CUB & Car  & Air      \\ \hline
MobileNet-v2   & 81.5       &  90.9     & 88.9\\
+RMG-PMSI (ours) & \textbf{87.0 (+5.5)} &\textbf{94.0 (+3.1)}&\textbf{91.6 (+2.7)} \\\hline
ResNet-18   & 82.0 & 90.5  & 87.5  \\
+RMG-PMSI (ours) & \textbf{86.8 (+4.8)} & \textbf{93.4 (+2.9)} &\textbf{90.2 (+2.7)} \\ \hline

ResNet-34  &   84.1      &     91.8      & 89.7  \\
+RMG-PMSI (ours)  &      \textbf{88.0 (+3.9)}       & \textbf{94.2 (+2.4)}   & \textbf{91.7 (+2.0)}               \\ \hline \hline
\end{tabular}}
\end{table}

\begin{table*}[]
\caption{Comparison of anti-interference ability of baseline and RMG-PMSI.}
\centering
\label{table:7}
\begin{tabular}{ccc|cc|cc}
\hline
                 & \multicolumn{2}{c}{CUB}              & \multicolumn{2}{c}{Car}             & \multicolumn{2}{c}{Air}               \\ \hline
                 & Baseline     & +RMG-PMSI             & Baseline     & +RMG-PMSI             & Baseline     & +RMG-PMSI              \\
Origin           & 81.5         & \textbf{87.0}           & 90.9         & \textbf{94.0}           & 88.9         & \textbf{91.6}          \\  \hline
+ Color-Jitter   & 14.1 (-77.4) & \textbf{59.9 (-27.1)} & 69.7 (-21.2) & \textbf{79.3 (-14.7)} & 64.7 (-24.2) & \textbf{73.9 ( -17.7)} \\
+ Gaussian-Noise & 22.2 (-59.3) & \textbf{78.6 (-18.4)} & 75.0 (-15.9) & \textbf{86.5 (-7.5)}  & 82.6 (-6.3)  & \textbf{86.6 (-5.0)}   \\ \hline
\end{tabular}
\end{table*}

\subsection{Robustness analysis}

\textbf{Stronger backbones.} In order to demonstrate the robustness and transferability of our proposed approach on different scales, in addition to MobilenetV2, we applied RMG-PMSI on ResNet-18 and ResNet-34 as well. For RMG-PMSI, we used Mix Acc (\%) as the evaluation metrics. The experimental results are shown in~\cref{table:6}.

On the three backbones, RMG-PMSI has a significant performance improvement compared to the baseline, which shows that RMG-PMSI is extremely effective on backbones of different scales. At the same time, compared to ResNet-18 and ResNet-34, RMG-PMSI brings more significant scores on MobilenetV2, implying that RMG-PMSI is more suitable for lightweight mobile networks.

\textbf{Anti-interference analysis.} After justifying the robustness of the RMG-PMSI on models of different scales. We further test the anti-interference capabilities of RMG-PMSI. We used Color-Jitter to generate interference data on the testset of the three datasets. We set the Jitter coefficient to be 1, i.e., the image's brightness, contrast, and saturation will be randomly adjusted to 0\% to 200\% of the original image. In addition, we also generated interference images with Gaussian-Noise (mean=0, variance=0, amplitude=5) on the testset of the three datasets. And then we test these two interference data respectively on the standard MobilenetV2 (baseline) and MobilenetV2 with RMG-PMSI. For RMG-PMSI, we used Mix Acc (\%) as the evaluation metrics and the experimental results are shown in~\cref{table:7}. It can be seen that the accuracy degradation of RMG-PMSI on the two types of interference data on the three testsets is less than the baseline. Especially on CUB200, the anti-interference ability of RMG-PMSI is far better than baseline. This may be because the distinction of birds mainly comes from some important local parts, such as eyes, feathers, and beaks. Adding interference will have a greater impact on global information, and the introduction of local information through RMG-PMSI can effectively alleviate this global interference.

\subsection{Visualization}
\begin{figure}
  \centering
   \includegraphics[width=1\linewidth]{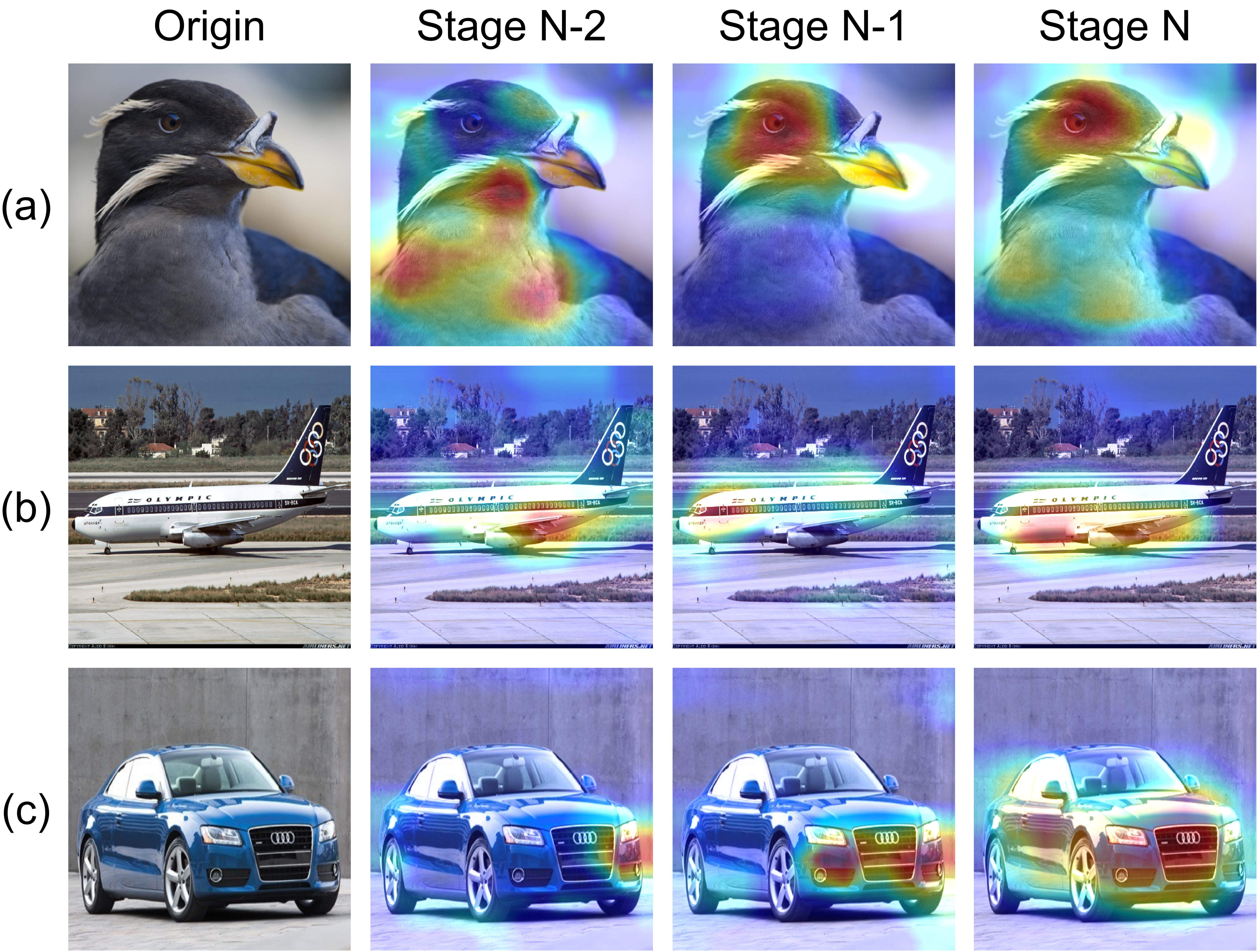}
   \caption{Activation map based on the MobilenetV2 with RMG-PMSI. Columns (a)-(c) are CUB, Air, and Car respectively.}
   \label{fig:heatmap}
\end{figure}

In order to demonstrate more insights into our approach, we applied
GradCAM \cite{DBLP:conf/iccv/SelvarajuCDVPB17} to visualize the convolutional layers of the last three stages of our method. The visualization results in~\cref{fig:heatmap} show that RMG-PMSI can help the model gradually shift from a fine-grained local feature to a large-grained global feature. For example, in bird pictures(a), the focus is on multiple local features on feathers in the shallow layer (Stage N-2) of the model. In the deeper layer (Stage N), besides focusing on the eye, which is the most distinguishing part, the model also pays attention to feathers, beaks, and other features that play a supporting role in classification. The visualization demonstrates that RMG-PMSI helps the model capture more distinguishing features from both the local and global perspectives, enhancing the performance and anti-interference of the model.

\section{Conclusion}
In this paper, we propose a Progressive Multi-Stage Interactive training method together with a Recursive Mosaic Generator (RMG-PMSI) for mobile networks of Fine-grained Visual Classification (FGVC) . By progressively capturing local to global features, RMG-PMSI effectively helps the model identify and integrate more distinguishing features and makes better use of the features extracted by mobile networks. Experiments on three prestigious fine-grained benchmarks prove that our method can significantly improve the performance compared to state-of-the-art lightweight models, with good robustness and transferability. For the future work, RMG-PMSI can incorporate more tricks such as attention mechanism.

{\small
\bibliographystyle{ieee_fullname}
\bibliography{report_template}

\begin{thebibliography}{10}\itemsep=-1pt

\bibitem{DBLP:conf/cvpr/AhnKS18}
Namhyuk Ahn, Byungkon Kang, and Kyung{-}Ah Sohn.
\newblock Image super-resolution via progressive cascading residual network.
\newblock In {\em {CVPR} Workshops}, pages 791--799. Computer Vision Foundation
  / {IEEE} Computer Society, 2018.

\bibitem{DBLP:conf/cvpr/BergB13}
Thomas Berg and Peter~N. Belhumeur.
\newblock {POOF:} part-based one-vs.-one features for fine-grained
  categorization, face verification, and attribute estimation.
\newblock In {\em {CVPR}}, pages 955--962. {IEEE} Computer Society, 2013.

\bibitem{DBLP:journals/corr/abs-2004-10934}
Alexey Bochkovskiy, Chien{-}Yao Wang, and Hong{-}Yuan~Mark Liao.
\newblock Yolov4: Optimal speed and accuracy of object detection.
\newblock {\em CoRR}, abs/2004.10934, 2020.

\bibitem{DBLP:journals/corr/BransonHBP14}
Steve Branson, Grant~Van Horn, Serge~J. Belongie, and Pietro Perona.
\newblock Bird species categorization using pose normalized deep convolutional
  nets.
\newblock {\em CoRR}, abs/1406.2952, 2014.

\bibitem{DBLP:conf/cvpr/ChenBZM19}
Yue Chen, Yalong Bai, Wei Zhang, and Tao Mei.
\newblock Destruction and construction learning for fine-grained image
  recognition.
\newblock In {\em {CVPR}}, pages 5157--5166. Computer Vision Foundation /
  {IEEE}, 2019.

\bibitem{DBLP:journals/corr/ClevertUH15}
Djork{-}Arn{\'{e}} Clevert, Thomas Unterthiner, and Sepp Hochreiter.
\newblock Fast and accurate deep network learning by exponential linear units
  (elus).
\newblock In {\em {ICLR} (Poster)}, 2016.

\bibitem{DBLP:conf/iclr/DosovitskiyB0WZ21}
Alexey Dosovitskiy, Lucas Beyer, Alexander Kolesnikov, Dirk Weissenborn,
  Xiaohua Zhai, Thomas Unterthiner, Mostafa Dehghani, Matthias Minderer, Georg
  Heigold, Sylvain Gelly, Jakob Uszkoreit, and Neil Houlsby.
\newblock An image is worth 16x16 words: Transformers for image recognition at
  scale.
\newblock In {\em {ICLR}}. OpenReview.net, 2021.

\bibitem{DBLP:conf/eccv/DuCBXMSG20}
Ruoyi Du, Dongliang Chang, Ayan~Kumar Bhunia, Jiyang Xie, Zhanyu Ma, Yi{-}Zhe
  Song, and Jun Guo.
\newblock Fine-grained visual classification via progressive multi-granularity
  training of jigsaw patches.
\newblock In {\em {ECCV} {(20)}}, volume 12365 of {\em Lecture Notes in
  Computer Science}, pages 153--168. Springer, 2020.

\bibitem{DBLP:conf/eccv/DubeyGGRFN18}
Abhimanyu Dubey, Otkrist Gupta, Pei Guo, Ramesh Raskar, Ryan Farrell, and
  Nikhil Naik.
\newblock Pairwise confusion for fine-grained visual classification.
\newblock In {\em {ECCV} {(12)}}, volume 11216 of {\em Lecture Notes in
  Computer Science}, pages 71--88. Springer, 2018.

\bibitem{DBLP:conf/cvpr/FuZM17}
Jianlong Fu, Heliang Zheng, and Tao Mei.
\newblock Look closer to see better: Recurrent attention convolutional neural
  network for fine-grained image recognition.
\newblock In {\em {CVPR}}, pages 4476--4484. {IEEE} Computer Society, 2017.

\bibitem{DBLP:conf/cvpr/GeLY19}
Weifeng Ge, Xiangru Lin, and Yizhou Yu.
\newblock Weakly supervised complementary parts models for fine-grained image
  classification from the bottom up.
\newblock In {\em {CVPR}}, pages 3034--3043. Computer Vision Foundation /
  {IEEE}, 2019.

\bibitem{DBLP:journals/corr/abs-2103-07976}
Ju He, Jieneng Chen, Shuai Liu, Adam Kortylewski, Cheng Yang, Yutong Bai,
  Changhu Wang, and Alan~L. Yuille.
\newblock Transfg: {A} transformer architecture for fine-grained recognition.
\newblock {\em CoRR}, abs/2103.07976, 2021.

\bibitem{DBLP:conf/cvpr/HeZRS16}
Kaiming He, Xiangyu Zhang, Shaoqing Ren, and Jian Sun.
\newblock Deep residual learning for image recognition.
\newblock In {\em {CVPR}}, pages 770--778. {IEEE} Computer Society, 2016.

\bibitem{DBLP:conf/cvpr/HuangLMW17}
Gao Huang, Zhuang Liu, Laurens van~der Maaten, and Kilian~Q. Weinberger.
\newblock Densely connected convolutional networks.
\newblock In {\em 2017 {IEEE} Conference on Computer Vision and Pattern
  Recognition, {CVPR} 2017, Honolulu, HI, USA, July 21-26, 2017}, pages
  2261--2269. {IEEE} Computer Society, 2017.

\bibitem{DBLP:conf/aaai/HuangWT21}
Shaoli Huang, Xinchao Wang, and Dacheng Tao.
\newblock Snapmix: Semantically proportional mixing for augmenting fine-grained
  data.
\newblock In {\em {AAAI}}, pages 1628--1636. {AAAI} Press, 2021.

\bibitem{DBLP:conf/cvpr/HuangXTZ16}
Shaoli Huang, Zhe Xu, Dacheng Tao, and Ya Zhang.
\newblock Part-stacked {CNN} for fine-grained visual categorization.
\newblock In {\em {CVPR}}, pages 1173--1182. {IEEE} Computer Society, 2016.

\bibitem{DBLP:conf/icml/IoffeS15}
Sergey Ioffe and Christian Szegedy.
\newblock Batch normalization: Accelerating deep network training by reducing
  internal covariate shift.
\newblock In {\em {ICML}}, volume~37 of {\em {JMLR} Workshop and Conference
  Proceedings}, pages 448--456. JMLR.org, 2015.

\bibitem{DBLP:conf/cvpr/KarrasLA19}
Tero Karras, Samuli Laine, and Timo Aila.
\newblock A style-based generator architecture for generative adversarial
  networks.
\newblock In {\em {CVPR}}, pages 4401--4410. Computer Vision Foundation /
  {IEEE}, 2019.

\bibitem{DBLP:conf/iccvw/Krause0DF13}
Jonathan Krause, Michael Stark, Jia Deng, and Li Fei{-}Fei.
\newblock 3d object representations for fine-grained categorization.
\newblock In {\em {ICCV} Workshops}, pages 554--561. {IEEE} Computer Society,
  2013.

\bibitem{DBLP:journals/tvt/LiYCMC19}
Xiaoxu Li, Liyun Yu, Dongliang Chang, Zhanyu Ma, and Jie Cao.
\newblock Dual cross-entropy loss for small-sample fine-grained vehicle
  classification.
\newblock {\em {IEEE} Trans. Veh. Technol.}, 68(5):4204--4212, 2019.

\bibitem{DBLP:journals/corr/abs-1712-00960}
Zuoxin Li and Fuqiang Zhou.
\newblock {FSSD:} feature fusion single shot multibox detector.
\newblock {\em CoRR}, abs/1712.00960, 2017.

\bibitem{DBLP:conf/cvpr/LinDGHHB17}
Tsung{-}Yi Lin, Piotr Doll{\'{a}}r, Ross~B. Girshick, Kaiming He, Bharath
  Hariharan, and Serge~J. Belongie.
\newblock Feature pyramid networks for object detection.
\newblock In {\em {CVPR}}, pages 936--944. {IEEE} Computer Society, 2017.

\bibitem{DBLP:conf/iccv/LinRM15}
Tsung{-}Yu Lin, Aruni RoyChowdhury, and Subhransu Maji.
\newblock Bilinear cnn models for fine-grained visual recognition.
\newblock In {\em {ICCV}}, pages 1449--1457. {IEEE} Computer Society, 2015.

\bibitem{DBLP:journals/corr/MajiRKBV13}
Subhransu Maji, Esa Rahtu, Juho Kannala, Matthew~B. Blaschko, and Andrea
  Vedaldi.
\newblock Fine-grained visual classification of aircraft.
\newblock {\em CoRR}, abs/1306.5151, 2013.

\bibitem{DBLP:conf/cvpr/RedmonF17}
Joseph Redmon and Ali Farhadi.
\newblock {YOLO9000:} better, faster, stronger.
\newblock In {\em {CVPR}}, pages 6517--6525. {IEEE} Computer Society, 2017.

\bibitem{DBLP:conf/cvpr/SandlerHZZC18}
Mark Sandler, Andrew~G. Howard, Menglong Zhu, Andrey Zhmoginov, and
  Liang{-}Chieh Chen.
\newblock Mobilenetv2: Inverted residuals and linear bottlenecks.
\newblock In {\em {CVPR}}, pages 4510--4520. Computer Vision Foundation /
  {IEEE} Computer Society, 2018.

\bibitem{DBLP:conf/iccv/SelvarajuCDVPB17}
Ramprasaath~R. Selvaraju, Michael Cogswell, Abhishek Das, Ramakrishna Vedantam,
  Devi Parikh, and Dhruv Batra.
\newblock Grad-cam: Visual explanations from deep networks via gradient-based
  localization.
\newblock In {\em {ICCV}}, pages 618--626. {IEEE} Computer Society, 2017.

\bibitem{DBLP:conf/iccv/ShahamDM19}
Tamar~Rott Shaham, Tali Dekel, and Tomer Michaeli.
\newblock Singan: Learning a generative model from a single natural image.
\newblock In {\em {ICCV}}, pages 4569--4579. {IEEE}, 2019.

\bibitem{DBLP:journals/corr/SimonyanZ14a}
Karen Simonyan and Andrew Zisserman.
\newblock Very deep convolutional networks for large-scale image recognition.
\newblock In {\em {ICLR}}, 2015.

\bibitem{WahCUB_200_2011}
C. Wah, S. Branson, P. Welinder, P. Perona, and S. Belongie.
\newblock The caltech-ucsd birds-200-2011 dataset.
\newblock Technical Report CNS-TR-2011-001, California Institute of Technology,
  2011.

\bibitem{DBLP:conf/cvpr/WeiXRXS00Y19}
Chen Wei, Lingxi Xie, Xutong Ren, Yingda Xia, Chi Su, Jiaying Liu, Qi Tian, and
  Alan~L. Yuille.
\newblock Iterative reorganization with weak spatial constraints: Solving
  arbitrary jigsaw puzzles for unsupervised representation learning.
\newblock In {\em {CVPR}}, pages 1910--1919. Computer Vision Foundation /
  {IEEE}, 2019.

\bibitem{DBLP:conf/iccv/WeiY0DL19}
Kun Wei, Muli Yang, Hao Wang, Cheng Deng, and Xianglong Liu.
\newblock Adversarial fine-grained composition learning for unseen
  attribute-object recognition.
\newblock In {\em {ICCV}}, pages 3740--3748. {IEEE}, 2019.

\bibitem{DBLP:journals/corr/WeiXW16}
Xiu{-}Shen Wei, Chen{-}Wei Xie, and Jianxin Wu.
\newblock Mask-cnn: Localizing parts and selecting descriptors for fine-grained
  image recognition.
\newblock {\em CoRR}, abs/1605.06878, 2016.

\bibitem{DBLP:conf/iccv/XieTHYZ13}
Lingxi Xie, Qi Tian, Richang Hong, Shuicheng Yan, and Bo Zhang.
\newblock Hierarchical part matching for fine-grained visual categorization.
\newblock In {\em {ICCV}}, pages 1641--1648. {IEEE} Computer Society, 2013.

\bibitem{DBLP:conf/cvpr/XieGDTH17}
Saining Xie, Ross~B. Girshick, Piotr Doll{\'{a}}r, Zhuowen Tu, and Kaiming He.
\newblock Aggregated residual transformations for deep neural networks.
\newblock In {\em 2017 {IEEE} Conference on Computer Vision and Pattern
  Recognition, {CVPR} 2017, Honolulu, HI, USA, July 21-26, 2017}, pages
  5987--5995. {IEEE} Computer Society, 2017.

\bibitem{DBLP:conf/eccv/YangLWHGW18}
Ze Yang, Tiange Luo, Dong Wang, Zhiqiang Hu, Jun Gao, and Liwei Wang.
\newblock Learning to navigate for fine-grained classification.
\newblock In {\em {ECCV} {(14)}}, volume 11218 of {\em Lecture Notes in
  Computer Science}, pages 438--454. Springer, 2018.

\bibitem{DBLP:conf/eccv/YuZZZY18}
Chaojian Yu, Xinyi Zhao, Qi Zheng, Peng Zhang, and Xinge You.
\newblock Hierarchical bilinear pooling for fine-grained visual recognition.
\newblock In {\em {ECCV} {(16)}}, volume 11220 of {\em Lecture Notes in
  Computer Science}, pages 595--610. Springer, 2018.

\bibitem{DBLP:conf/iccv/YunHCOYC19}
Sangdoo Yun, Dongyoon Han, Sanghyuk Chun, Seong~Joon Oh, Youngjoon Yoo, and
  Junsuk Choe.
\newblock Cutmix: Regularization strategy to train strong classifiers with
  localizable features.
\newblock In {\em {ICCV}}, pages 6022--6031. {IEEE}, 2019.

\bibitem{DBLP:conf/iclr/ZhangCDL18}
Hongyi Zhang, Moustapha Ciss{\'{e}}, Yann~N. Dauphin, and David Lopez{-}Paz.
\newblock mixup: Beyond empirical risk minimization.
\newblock In {\em {ICLR} (Poster)}. OpenReview.net, 2018.

\bibitem{DBLP:conf/eccv/ZhangDGD14}
Ning Zhang, Jeff Donahue, Ross~B. Girshick, and Trevor Darrell.
\newblock Part-based r-cnns for fine-grained category detection.
\newblock In {\em {ECCV} {(1)}}, volume 8689 of {\em Lecture Notes in Computer
  Science}, pages 834--849. Springer, 2014.

\bibitem{DBLP:conf/iccv/ZhengFML17}
Heliang Zheng, Jianlong Fu, Tao Mei, and Jiebo Luo.
\newblock Learning multi-attention convolutional neural network for
  fine-grained image recognition.
\newblock In {\em {ICCV}}, pages 5219--5227. {IEEE} Computer Society, 2017.

\bibitem{DBLP:conf/icmcs/ZhengC0M20}
Yixiao Zheng, Dongliang Chang, Jiyang Xie, and Zhanyu Ma.
\newblock Iu-module: Intersection and union module for fine-grained visual
  classification.
\newblock In {\em {ICME}}, pages 1--6. {IEEE}, 2020.

\bibitem{DBLP:conf/eccv/ZhouHCFY20}
Daquan Zhou, Qibin Hou, Yunpeng Chen, Jiashi Feng, and Shuicheng Yan.
\newblock Rethinking bottleneck structure for efficient mobile network design.
\newblock In {\em {ECCV} {(3)}}, volume 12348 of {\em Lecture Notes in Computer
  Science}, pages 680--697. Springer, 2020.

\bibitem{DBLP:conf/aaai/ZhuangW020}
Peiqin Zhuang, Yali Wang, and Yu Qiao.
\newblock Learning attentive pairwise interaction for fine-grained
  classification.
\newblock In {\em {AAAI}}, pages 13130--13137. {AAAI} Press, 2020.

\end{thebibliography}
}

\end{document}